\documentclass[10pt,twocolumn,letterpaper]{article}

\usepackage{algorithm}
\usepackage{algorithmic}
\usepackage{amsmath}
\usepackage{multirow}

\usepackage[pagenumbers]{cvpr} % To force page numbers, e.g. for an arXiv version

\definecolor{cvprblue}{rgb}{0.21,0.49,0.74}
\usepackage[pagebackref,breaklinks,colorlinks,allcolors=cvprblue]{hyperref}

\title{UVCG: Leveraging Temporal Consistency for Universal Video Protection}

\author{KaiZhou Li\\
Tsinghua University\\
China\\
{\tt\small lkz23@mails.tsinghua.edu.cn}
\and
Jindong Gu\\
University of Oxford\\
UK\\
{\tt\small jindong.gu@eng.ox.ac.uk}
\and
Xinchun Yu\\
Tsinghua University\\
China\\
{\tt\small yuxinchun@sz.tsinghua.edu.cn}
\and
Junjie Cao\\
Tsinghua University\\
China\\
{\tt\small cjj23@mails.tsinghua.edu.cn}
\and
Yansong Tang\\
Tsinghua University\\
China\\
{\tt\small tang.yansong@sz.tsinghua.edu.cn}
\and
Xiao-Ping Zhang\textsuperscript{*}\\
Tsinghua University\\
China\\
{\tt\small xpzhang@ieee.org}
}

\begin{document}
\maketitle
\begin{abstract}
The security risks of AI-driven video editing have garnered significant attention. Although recent studies indicate that adding perturbations to images can protect them from malicious edits, directly applying image-based methods to perturb each frame in a video becomes ineffective, as video editing techniques leverage the consistency of inter-frame information to restore individually perturbed content. To address this challenge, we leverage the temporal consistency of video content to propose a straightforward and efficient, yet highly effective and broadly applicable approach, Universal Video Consistency Guard (UVCG). UVCG embeds the content of another video(target video) within a protected video by introducing continuous, imperceptible perturbations which has the ability to force the encoder of editing models to map continuous inputs to misaligned continuous outputs, thereby inhibiting the generation of videos consistent with the intended textual prompts. Additionally leveraging similarity in perturbations between adjacent frames, we improve the computational efficiency of perturbation generation by employing a perturbation-reuse strategy. We applied UVCG across various versions of Latent Diffusion Models (LDM) and assessed its effectiveness and generalizability across multiple LDM-based editing pipelines. The results confirm the effectiveness, transferability, and efficiency of our approach in safeguarding video content from unauthorized modifications. 
\end{abstract}    
\section{Introduction}
\label{sec:intro}

Generative technology has attracted considerable attention, especially in image processing, prompting the development of various image generation models\cite{goodfellow2014generative}\cite{DiffusionModel}\cite{vae}\cite{LDM}. Among these, Latent Diffusion Models (LDM) \cite{LDM} have recently demonstrated notable success in image editing tasks\cite{bodur2024iedit}\cite{brooks2023instructpix2pix}\cite{zhang2023sine}. Given LDM's superior performance in this domain, several studies have extended its applicability from 2D image editing to the spatiotemporal video editing field\cite{gen-1}\cite{video-p2p}\cite{tuneavideo}. State-of-the-art open-source video editing models\cite{tokenflow}\cite{text2video}\cite{tuneavideo} now allow for the production of realistic, edited videos through the input of concise textual prompts. The emergence of advanced video editing models has made video editing far more efficient and accessible, which was previously a highly demanding task requiring extensive manual effort. While this accessibility enhances user convenience, it also amplifies significant security concerns \cite{gu2024responsible}, including the potential misuse by malicious actors to manipulate sensitive or harmful video content\cite{he2024diff}\cite{yu2024cross}. This could involve the fabrication of videos featuring both public figures and private individuals or the creation of videos intended to deceive, intimidate, or manipulate emotions. Since video serves as a critical source of information, failure to ensure its authenticity exacerbates the challenges of verifying information reliability, posing risks to public trust and security.

\begin{figure}[t]
  \centering
   \includegraphics[width=0.9\linewidth]{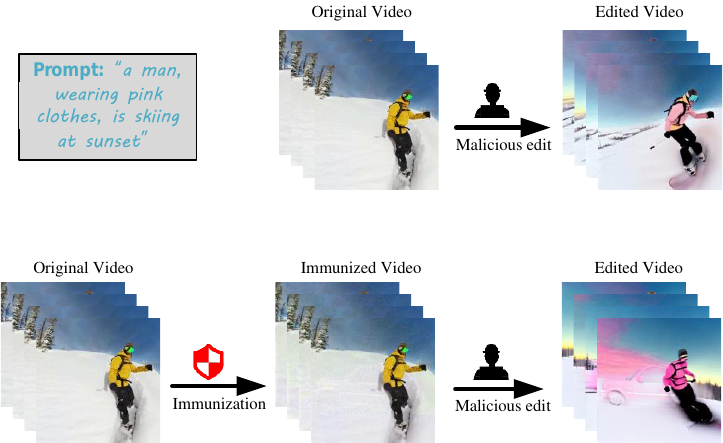}
   \caption{Overview of Framework. An attacker can modify the content of a video according to their intent by using textual descriptions and then generate malicious videos through any video editing pipeline (top). We can immunize the video by introducing imperceptible perturbations, thereby disrupting their ability to perform such edits (bottom).}
   \label{fig:overview}
\end{figure}

Previous research has demonstrated that introducing carefully crafted  perturbations into images can effectively prevent malicious edits or unauthorized use for artistic style transfer\cite{diffattack}\cite{zheng2023understanding}\cite{mist}\cite{liang2023adversarial}. Similar to image generation protection, an overview of video editing protection is presented in Figure \ref{fig:overview}. However, research focused on video protection remains limited. An intuitive approach is to adapt image-based protection methods for videos. Unlike images, videos inherently exhibit temporal continuity, with sequential frames combined to convey motion information. To maintain temporal coherence in edited videos, video editing pipelines typically integrate large-scale diffusion models with specialized editing techniques. For instance, Tune-A-Video, proposed by Wu \etal \cite{tuneavideo}, extends the U-Net architecture within LDM into 3D to ensure temporal consistency in the edited content, while TokenFlow \cite{tokenflow} leverages feature propagation to maintain coherence in feature dynamics. Independently applying image-based protection methods to each frame of video neglects the temporal continuity between frames and the error-correction capabilities introduced by advanced editing method. Meanwhile, this approach results in significant computational costs. Moreover, the variety of editing method significantly complicates efforts to safeguard video content, as it becomes difficult to predict which specific editing method and which versions of LDM a malicious actor might employ for unauthorized alterations. Therefore, effective video protection must account for both the characteristics of videos, the efficiency of the immunization process, and the generalizability of the immunization\cite{liu2016delving}\cite{papernot2016transferability}.

In this work, we leverage the consistency of video content to propose a straightforward yet universal video editing protection method—Universal Video Consistency Guard (UVCG), which requires minimal GPU time to generate immunization while achieving superior effectiveness and generalizability. Unlike image-based methods, our approach ensures the consistency of perturbed videos by selecting a target video to guide the perturbation optimization, where we employ projected gradient descent(PGD)\cite{goodfellow2014explaining}\cite{pgd} to solve the optimization problem. These perturbations force the encoder to map continuous inputs to misaligned continuous outputs, preventing the editing pipeline from correcting distortions through inter-frame consistency. Meanwhile, as the perturbed video still preserves temporal consistency, the immunization remains effective across various editing pipelines built on different LDM versions. Additionally, we observed that the perturbation between adjacent frames exhibits a certain degree of similarity. Therefore, to enhance the efficiency of our approach, we employ a perturbation-reuse strategy to reduce the optimization time needed for perturbation generation\cite{liao2023inter}. Given that the choice of target video impacts the effectiveness of immunization, we summarize two optimal target selection strategies based on our experimental findings.

To evaluate the performance of our proposed solution, we applied perturbations to videos using two different versions of LDM and evaluated the effectiveness and generalizability of our immunization across various LDM-based editing pipelines, including TokenFlow\cite{tokenflow}, Text2Video-Zero\cite{text2video}, Tune-A-Video\cite{tuneavideo}, and FateZero\cite{fatezero}. The experimental results demonstrate that, regardless of whether the same LDM version was used for both immunization and editing, the edited videos exhibited significant distortions. In TokenFlow, the alignment between video content and text descriptions dropped from 0.32 to 0.28 and 0.30. Compared to the baseline method with random noise, the similarity in Text2Video-Zero decreased from 71.94 to 32.20 and 34.72 after immunization. Additionally, our method consumes relatively low computational resources and achieves a protection success rate of 87\% in human evaluations. These results confirm the effectiveness, generalizability and efficient of our approach. Our work makes several key contributions:
 \begin{itemize}
     \itemsep0em
     \item We propose a method to safeguard videos from malicious editing—Universal Video Consistency Guard (UVCG). By leveraging the temporal consistency of video content, our approach achieves effective immunization without being affected by the complexity and variability of video editing pipelines, while possessing generalizability across different LDM versions.
    \item We employ a noise-reuse strategy to improve the efficiency of the protection process and provide guidelines for selecting target videos to enhance immunization effectiveness.
    \item We conducted extensive qualitative and quantitative experiments, along with survey-based evaluations, to assess UVCG. The results demonstrate the effectiveness, generalizability, and efficiency of our method in protecting video from being maliciously edited.
 \end{itemize}

%-------------------------------------------------------------------------

% \input{sec/2_Preliminaries}
\section{Related work}
\label{sec:relwork}
\subsection{Projection gradient descent}
Projection Gradient Descent (PGD) is a widely used technique for generating adversarial examples. These attacks involve introducing small perturbations to input data, such as images, which remain imperceptible to human observers but cause a trained model to produce incorrect predictions. Let \( \mathbf{x} \) denote the original input and \( \mathbf{y} \) the corresponding true label. Adversarial attacks aim to find a perturbation \( \mathbf{\delta} \) such that the perturbed input \( \mathbf{x}' = \mathbf{x} + \mathbf{\delta} \) results in the model \( f_\theta(\mathbf{x}') \) misclassifying it, i.e., \( f_\theta(\mathbf{x}') \neq \mathbf{y} \). The perturbation \( \mathbf{\delta} \) is constrained to lie within an \( \ell_p \)-norm ball of radius \( \epsilon \), ensuring the perturbation is small enough to be imperceptible:\(\|\mathbf{\delta}\|_p \leq \epsilon\). PGD improves on this process by generating the adversarial perturbations iteratively. Starting from an initial perturbation \( \mathbf{\delta}_0 \), PGD updates the perturbation in the direction of the gradient of the loss function \( L(\mathbf{x}, \mathbf{y}, \theta) \) with respect to the input:
\begin{equation}
\mathbf{\delta}^{t+1} = \mathbf{\delta}^t + \alpha \cdot \text{sign} \left( \nabla_{\mathbf{x}} L(f_\theta(\mathbf{x}+\delta^t), \mathbf{y}) \right),
\label{eq:pgd}
\end{equation}
where \( \alpha \) is the step size. This gradient ascent step maximizes the loss, thereby increasing the likelihood of the model misclassifying the adversarial example.  This iterative process is repeated for steps \( T \), yielding the final adversarial perturbation \( \mathbf{\delta}^T \).

\subsection{Adversarial protections in image editing}
While generative models for image editing have significantly advanced creative work, they have also raised concerns about potential misuse for illegal purposes. The robust generative capabilities of Latent Diffusion Models (LDMs) have intensified worries regarding security implications. To mitigate these risks, researchers have investigated adversarial examples to protect images from unauthorized edits\cite{diffattack}\cite{photoguard}. Notably, Photoguard\cite{photoguard}, proposed by Salman \etal, has demonstrated remarkable effectiveness in protecting images from malicious manipulations by large-scale diffusion models. They introduced two strategies—encoder attack and diffusion attack—disrupting the editing process by forcing the encoder or the entire diffusion process to map the input images to some misaligned outputs\cite{photoguard}. Furthermore, some studies introduce perturbations into specific artworks to map the artist's style to other artistic styles, preventing the unauthorized use of their works for training style imitation models\cite{mist}\cite{liang2023adversarial}\cite{glaze}. However, protections specific to video editing remain limited. Therefore, in the context of LDMs, we leverage the consistency of video content to propose a straightforward yet universal video editing protection method.
\section{Method}
\label{sec:method}
In this section, we introduce the threat model and elaborate on the design concept and detailed implementation of the Universal Video Consistency Guard (UVCG). Additionally, we provide recommendations for selecting an appropriate target video for immunizing the video.
\subsection{Threat model}
\textbf{Attacker Capabilities and Intentions.} Attackers can access pre-trained open-source models from public platform and  select various editing pipelines based on their objectives. As models and editing techniques continue to advance, the resulting modifications become increasingly indistinguishable from authentic content. By exploiting the latest models and tools, attackers can maliciously alter videos, creating significant challenges for video protection.

\noindent\textbf{Defender Capabilities and Intentions.} Defenders also have access to open-source models available on public platforms, but they lack insight into the models versions and editing techniques employed by attackers. To maintain effective video protection, defenders must ensure that their methods are resilient over time and robust against various model versions and editing pipelines.

\subsection{Universal video consistency guard}
\label{sec:UVCG}
In the LDM model, for computational efficiency, an image $\mathbf{x}$ is transformed into a latent representation $\mathbf{z}$, which is then used to generate a new image( further details on LDM, refer to Appendix \ref{sec:Latent Diffusion Model}). Thus, the most straightforward approach is to modify these latent representation so that the model operates on the altered representation $\mathbf{z}'$. In previous work, PRIME\cite{prime} apply the image protection methods to each frame of the video and selects different target images for consecutive frames, resulting in inconsistent features between frames. However, When editing a specific video frame, the editing pipeline often applies global or local constraints to regulate the content, providing the model with "correction" capabilities. PRIME's disruption of video content consistency could be "corrected" by the global constraints within the editing pipeline.

\begin{figure*}[t]
  \centering
   \includegraphics[width=0.8\linewidth]{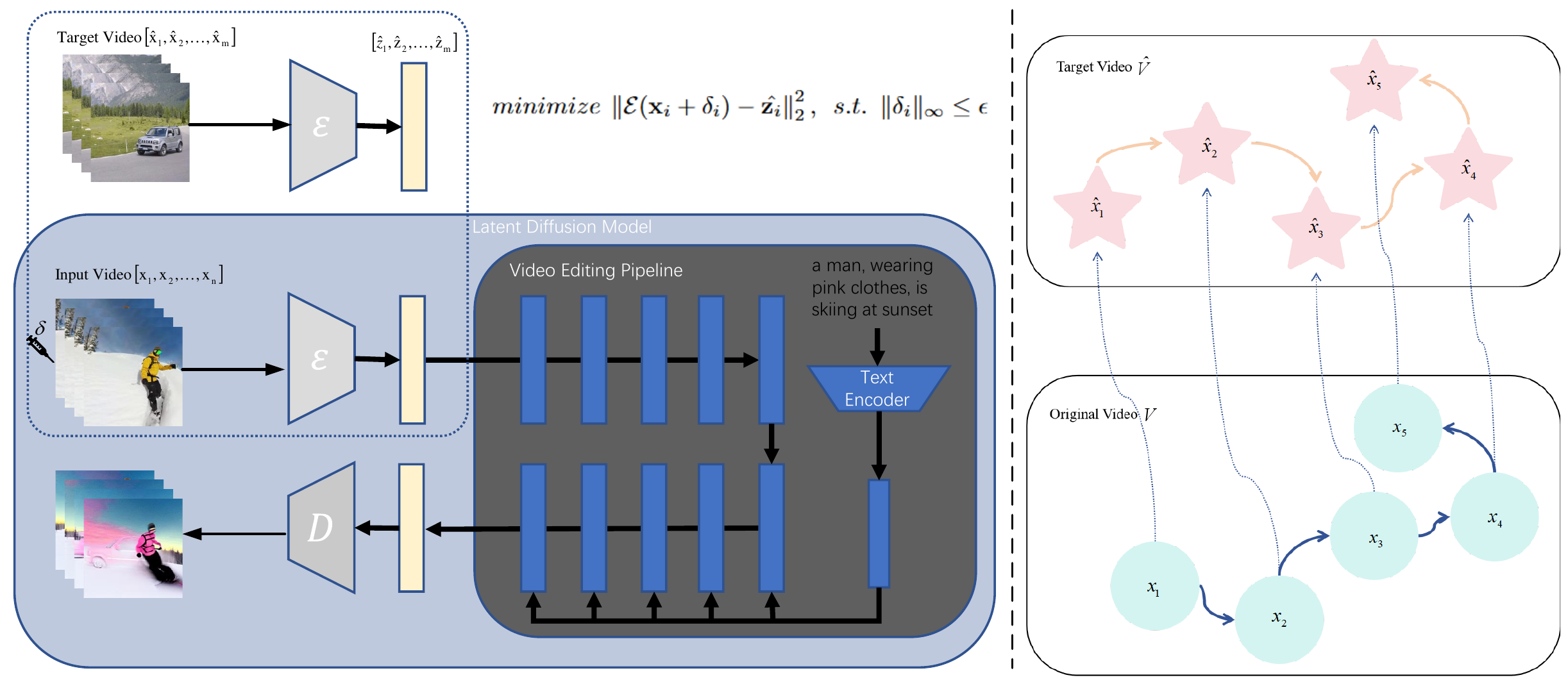}
   \caption{\textbf{Right:} Overview of UVCG. When applying UVCG, our goal is to map the continuous  representations of original video to the continuous representations of target video. \textbf{Left:} Feature Transfer. The top represents the feature space of the target video, while the bottom represents the feature space of the original video. By adding continuous perturbations, we guide the feature space of the original video towards that of the target video (indicated by the dashed line in the figure).}
   \label{fig:algorithm}
\end{figure*}

A more effective approach is to align the video with another continuous feature space. The key distinction between video editing and image editing lies in the need for video editing to capture not only the features of individual frames but also how these features evolve within the latent space (as indicated by the solid arrows in right-side of Figure \ref{fig:algorithm}). To prevent the model from correcting the alterations while compelling it to learn incorrect feature transitions, we introduce perturbations into video frames to map the original latent representations to another set of representations with consistent content. Left-side of Figure \ref{fig:algorithm} provides a schematic overview of our method. Specifically, we select a different video as the target and use the latent representations of the consecutive frames from this target video to guide the perturbation calculation. This method perturbs the original video's latent features to align with those of the target video(alignment process see in Figure \ref{fig:algorithm}), leading the editing pipeline recognizing a set of continuous but "incorrect" features.

Given a video \( V = [\mathbf{x}_1, \mathbf{x}_2, \ldots, \mathbf{x}_n] \) to be protected, a target video \( \hat{V} = [\hat{\mathbf{x}}_1, \hat{\mathbf{x}}_2, \ldots, \hat{\mathbf{x}}_m] \) is selected. Each frame of the target video is encoded into a latent vector representation \( \hat{\mathbf{z}}_i \) using the encoder \( \mathcal{E} \) resulting in a sequence of latent representations with consistent content: \( \hat{Z} = [\hat{\mathbf{z}}_1, \hat{\mathbf{z}}_2, \ldots, \hat{\mathbf{z}}_m] \). If the number of latent representations \( m \) in the target video is less than the number of frames in the protected video, the sequence \( \hat{Z} \) will be reused. To force the encoder to map continuous inputs to misaligned continuous outputs, we formulate the following optimization problem for each frame \(\mathbf{x}_i\):
\begin{equation}
  {minimize}\ \left\|\mathcal{E}({\mathbf{x}_i}+\delta_i)-\hat{\mathbf{z}_i}\right\|_2^2,\ \ s.t.\ \ \|\delta_i\|_\infty\leq\epsilon
  \label{eq:uvcg}
\end{equation}
where \( \epsilon \) is the range of the perturbation values. In our experiments, we use PGD to obtain an approximate solution for the perturbation in this optimization problem. Specifically, we use Equation \ref{eq:uvcg} as the loss function of Equation \ref{eq:pgd} and fix the number of PGD iterations to \(T\) for each frame. Finally we get a small imperceptible pertubation \(\delta_i^T\), which is applied to the \(i\)-th frame \(\mathbf{x}_i\) of the video. After perturbations are added to all frames, the resulting immunized video closely aligns with the target video within the latent feature space. Specifically, the consecutive frames of the immunized video are mapped sequentially onto the latent representation sequence \(\hat{Z}\) of the target video. Additionally, we observed that the added perturbations exhibit a certain degree of similarity across frames. To enhance the efficiency of immunization, we employ a perturbation reuse strategy, the initial perturbation \( \delta_i^0 \) initialized according to the following equation:

\begin{equation}
  \delta_i^0=
  \begin{cases}
    \delta_{unif}&i=0\\
    \delta_{i-1}^T&i=1,2,\ldots,m
  \end{cases}
  \label{eq:noise}
\end{equation}
where \(\delta_{unif} \sim \mathcal{U}(-\epsilon, \epsilon)\). The optimized perturbation from the previous frame \( \delta_{i-1}^T \) serves as the initial perturbation for the next frame \( i \). This approach accelerates the convergence of the optimization. The complete procedure for adding perturbations to the video is summarized in Algorithm \ref{alg:UVCG}.

\begin{algorithm}[!hbtp]    
\caption{Universal Video Consistency Guard}
\label{alg:UVCG}
    \begin{algorithmic}[1]
        \STATE {\bfseries Input:} Video \(V=[\mathbf{x}_1, \mathbf{x}_2,\ldots, \mathbf{x}_n]\), Target Video \(\hat{V}=[\hat{\mathbf{x}}_1, \hat{\mathbf{x}}_2, \ldots,\hat{\mathbf{x}}_m ]\), model \(\mathcal{E}\), perturbation budget \(\epsilon\), step size \(k\), number of PGD steps \(T\).
        \STATE \(V' \gets [\ ]\)
        \FOR{\(i = 1 \ldots n\)}
            \IF{\(i>0\)}
                \STATE \(\delta_{i}^0 \gets \delta_{i-1}^T\)
            \ELSE
                \STATE \(\delta_i^0 \gets \delta_{rand}\)
            \ENDIF
            \STATE Obtain target latent representation \(\hat{\mathbf{z}}_i \gets \mathcal{E}(\hat{\mathbf{x}}_i)\)
            \FOR{\(t = 0 \ldots T\)}
                \STATE \(\mathbf{x}_i'= \mathbf{x}_i+\delta_{i}^t\)
                \STATE Update \(\delta_{i}^t=clip(\delta_{i}^t+\alpha\cdot \text{sign}\left[\nabla_{\mathbf{x}} L (\mathcal{E}(\mathbf{x}_i'), \hat{\mathbf{z}}_i)\right])\)
            \ENDFOR
            \STATE Append \(\mathbf{x}_i'=\mathbf{x}_i+\delta_{i}^T\) to \(V'\)
        \ENDFOR
        \STATE {\bfseries Return:} \(V'\)
    \end{algorithmic}
\end{algorithm}

\subsection{Target video selection recommendations}
\label{sec:select_recommend}
As demonstrated in Equation \ref{eq:pgd}, we introduce the target video \(\hat{Z}\)-sequence as guiding variable for optimizing perturbations. Our observations suggest that the effectiveness of video protection is influenced by the choice of the target video. Selecting an appropriate target video can not only enhance the effectiveness of the immunity but also enhances its transferability across different editing models. 

Through experimentation, we recommend two strategies for selecting target videos that provide optimal protection. In image-based tasks, semantically similar images tend to cluster closely in feature space, and this principle applies equally to videos. Therefore, when the primary content of the target video falls within the same category as that of the protected video, it becomes easier to mislead the model’s recognition process, thereby enhancing the protective effect (see Figure \ref{fig:result_Fatezero_01} - Figure \ref{fig:results_text2video_03} in Appendix \ref{sec:addition Results}). 

The second strategy focuses on selecting videos with simple content and prominent subjects as target videos. Our experiments demonstrate that videos with clear focal elements and minimal complexity also provide robust protection (see Figure \ref{fig:results_tokenflow_01}, and the right-side of Figure \ref{fig:result_TuneAVideo_01}, the left-side of Figure \ref{fig:result_TuneAVideo_02} in Appendix \ref{sec:addition Results}).
\section{Experiment}
\subsection{Experimental settings}
\textbf{Dataset.} We selected 40 representative videos from the DAVIS\cite{davis} dataset, each with a spatial resolution of 512×512 pixels, and consisting of 8 to 70 frames. Automatic captions for the videos were generated using BLIP-2 \cite{blip}. Additionally, 80 editing prompts were designed for each editing model, tailored to their respective strengths in different editing types. These prompts encompass object editing, background modification, and style transformation.

\noindent \textbf{Immune Settings.} We assume that the attacker's editing pipeline generates high-quality, realistic videos using either the LDM model or a fine-tuned version of it. Regarding the editing pipeline parameters, they are adjusted based on the editing performance on the original video at first, and the same hyperparameters are then used for editing the immunized video. Moreover, we assume that the defender has no prior knowledge of the model version used by the attacker. Therefore, we implement our immunity approach using publicly available open-source models. Specifically, we utilize the two most popular open-source models: Stable Diffusion v1-4 and Stable Diffusion v2-1. Regarding the parameter choices for the PGD attack, the \( l_\infty \) norm is employed to constrain the perturbation magnitude \( \epsilon \) within 15/255. The optimization process involves 200 steps with a step size of 2/255.

% \noindent \textbf{Quantitative Metrics.} We use evaluation metrics commonly employed in video editing models\cite{tokenflow}\cite{text2video}\cite{tuneavideo}-prompt consistency and frame consistency—to assess video editing quality. Leveraging CLIP\cite{clip}, we compute visual embeddings for each frame of the output videos and measure both the average cosine similarity between consecutive frame pairs and the overall alignment between video content and editing prompts. Additionally, to quantify differences between protected and non-protected videos, we apply standard image similarity metrics, including SSIM\cite{ssim}, PSNR, and LPIPS\cite{lpips}, averaging these scores across all frame pairs for video pairs. We also compute the VMAF\cite{vmaf} score, a metric specifically developed for evaluating video quality. Finally, we assess the resource consumption involved in the immunization process.

\subsection{Qualitative results.}

\textbf{Protection Effectiveness.} We consider the protection successful under the following two scenarios. First, from the perspective of distinguishing the video as either genuine or generated, the protection is deemed effective if viewers can easily recognize the video as being artificially generated with minimal cognitive effort. Second, the protection is considered successful if the video lacks distinctive features aligned with the given editing prompts.

\begin{figure*}
  \centering
  \includegraphics[width=\linewidth]{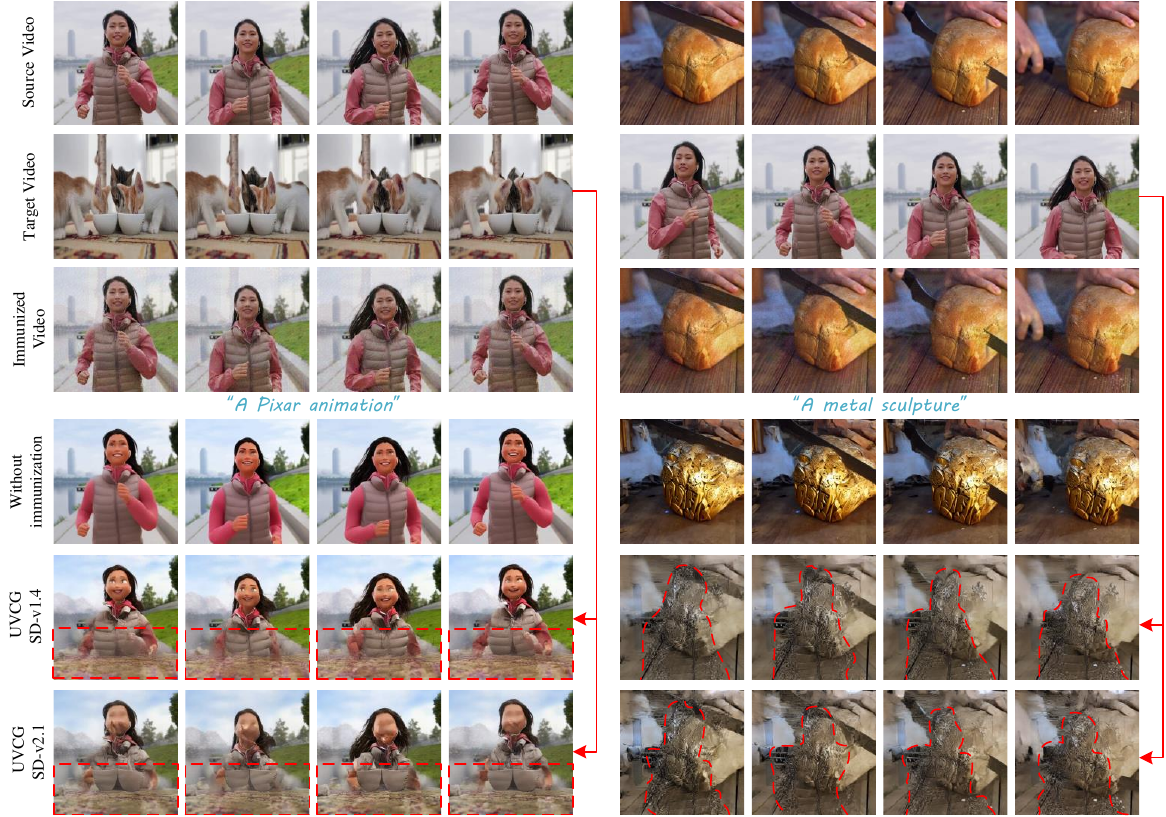}
  \caption{\textbf{Protection Effectiveness on Tokenflow.} The base model used for editing the video on the left is SD-v2.1, while the one used for the right-side video is SD-v1.5. First row: The original video. Second row: The target video. Third row: The immunized video. Fourth row: editing results without immunization. Fifth row: Editing results after applying UVCG with SD-v1.4 as the protection model. Sixth row: Editing results after UVCG using SD-v2.1 as the protection model.}
  \label{fig:results_tokenflow_01}
\end{figure*}

We first verify that the models are capable of generating videos that align with text commands. For style transfer, given a video of a woman running and the editing prompt "A Pixar animation". As expected, TokenFlow \cite{tokenflow} successfully generated a video that matches the description(see Figure \ref{fig:results_tokenflow_01}, second row on the left). For object editing, when provided with a video of a swan swimming by a river and the editing prompt, "replace swan with mallard," Text2Video-Zero \cite{text2video} also accurately performed the requested edit (see Figure \ref{fig:result_text2video_01}, second row on the left). Further examples of video editing can be found in the Appendix \ref{sec:addition Results}.

Secondly, we evaluate the performance of the protection method. When the models attempt to edit the immunized video, as discussed in Section \ref{sec:UVCG}, they fail to  capture the correct video features and instead learn the erroneous ones introduced by our perturbations. As various editing outcomes of TokenFlow show in Figure \ref{fig:results_tokenflow_01}, after applying our method to immunize the video, each frame of the generated video appear floor-like features in the lower half, a characteristic inherited from the target video we selected. Furthermore, due to the inclusion of other latent features, the expressions of the characters in the edited video appear distorted. In the right-side video, where the original left-side video is used as the target, following our target selection recommendations: the original left-side video has simple content and a clear focal point, serves as an ideal choice for the target. TokenFlow\cite{tokenflow} fails to alter the structure of the bread, and each frame of the video shows a noisy shape resembling a woman in the foreground. Figure \ref{fig:result_text2video_01} display the editing results on Text2Video-Zero\cite{text2video}. In the left-side video, a target video of a walking camel was selected, resulting in a noticeable camel artifact appearing throughout the edited video. The same protective effect can also be observed in the right-side video. The effectiveness of video protection across other editing pipelines is shown in Appendix \ref{sec:addition Results}.
\begin{figure*}
  \centering
  \includegraphics[width=\linewidth]{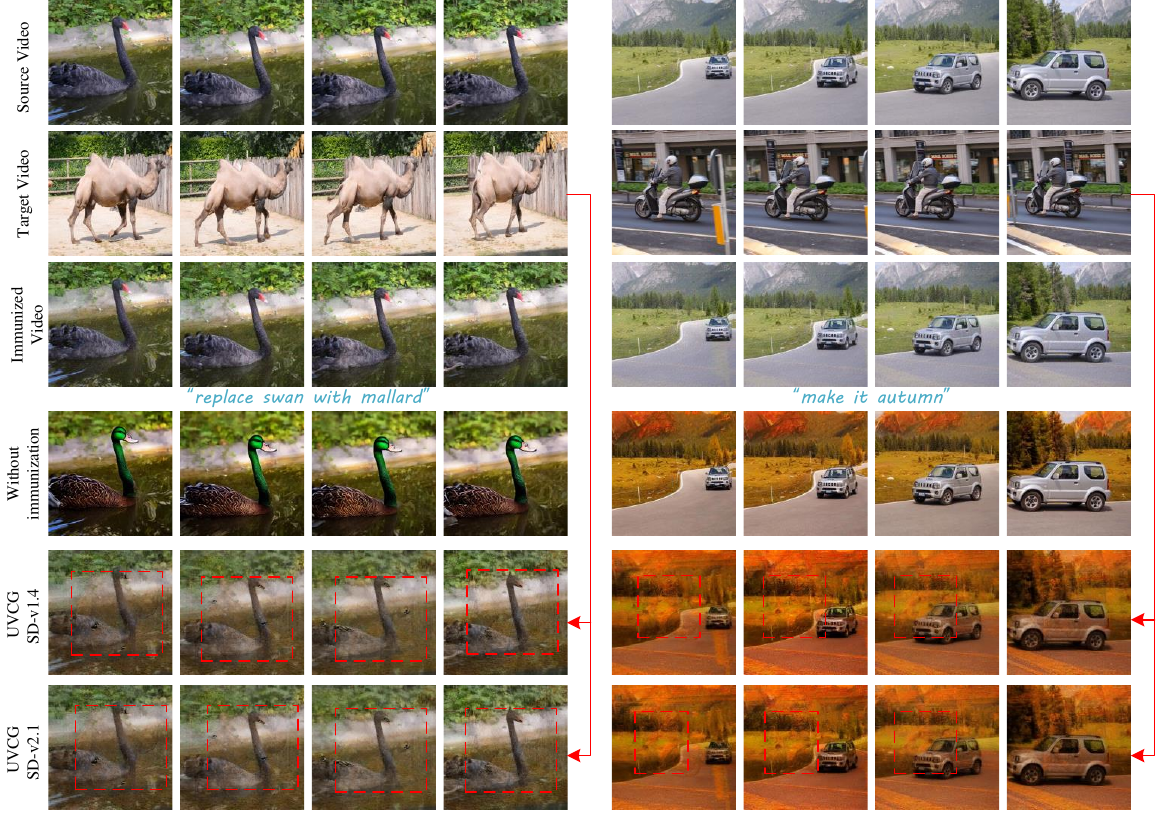}
  \caption{\textbf{Protection Effectiveness on Text2Video-zero.} The base model employed for editing in Text2Video-zero is Instruct-pix2pix\cite{instructpix2pix}. First row: The original video. Second row: The target video. Third row: The immunized video. Fourth row: editing results without immunization. Fifth row: Editing results after applying UVCG with SD-v1.4 as the protection model. Sixth row: Editing results after UVCG using SD-v2.1 as the protection model.}
  \label{fig:result_text2video_01}
\end{figure*}

\noindent \textbf{Generalization of Immunization.} From a defensive perspective, the transferability of immunization is crucial. With numerous video editing pipeline and LDM verisons available, tailoring defense mechanisms to each specific editing pipeline would would not achieve truly effective defense.  Therefore, immunization methods must demonstrate a certain degree of generalization. Figures \ref{fig:results_tokenflow_01} and \ref{fig:result_text2video_01} present the visual editing results of videos protected with different LDM versions across various editing pipelines. We could observe that when the base model used by the attacker matches the one employed for video immunization, the protective visual effect is more pronounced. Furthermore, even if the attacker’s model differs from the one used during immunization, our protection remains effective. This demonstrates that our immunization approach exhibits strong transferability across different versions of LDMs and various editing pipelines.

\subsection{Quantitative results}
We introduces uniform random noise as the baseline for comparison with our proposed immunity method, maintaining the same perturbation intensity as our approach. We use evaluation metrics commonly employed in video editing models\cite{tokenflow}\cite{text2video}\cite{tuneavideo}-prompt consistency and frame consistency—to assess video editing quality. Leveraging CLIP\cite{clip}, we compute visual embeddings for each frame of the output videos and measure both the average cosine similarity between consecutive frame pairs and the overall alignment between video content and editing prompts. Additionally, to quantify differences between protected and non-protected videos, we apply standard image similarity metrics, including SSIM\cite{ssim}, PSNR, and LPIPS\cite{lpips}, averaging these scores across all frame pairs for video pairs. We also compute the VMAF\cite{vmaf} score, a metric specifically developed for evaluating video quality. Finally, we assess the resource consumption involved in the immunization process.

\noindent \textbf{Consistency scores.} We report Consistency scores in Table \ref{tab:metrics}. The results indicate that applying random noise offers minimal protection, as the generated/edited videos are nearly identical to those produced without any immunity. In contrast, our proposed immunity method introduces additional features that disrupt the ability of editing models to recognize key video content characteristics, thereby reducing the alignment between the generated content and the editing descriptions. Additionally, using SD-v1.4 or SD-v2.1 as immunity models significantly decreases the alignment between the video content and the textual prompts demonstrating the robust generalizability of our method across different model versions.

\noindent \textbf{Similarity scores.} The similarity scores are summarized in Table \ref{tab:metrics}. The results indicate that, compared to the random noise method, our immunity method produces edits that differ significantly from those generated without immunity. Furthermore, regardless of which open-source model is employed for protection, the similarity between the immune-generated edits and the non-immune edits remains consistently low, further validating the robust generalizability of our method.

\begin{figure}
  \centering
  \includegraphics[width=\linewidth]{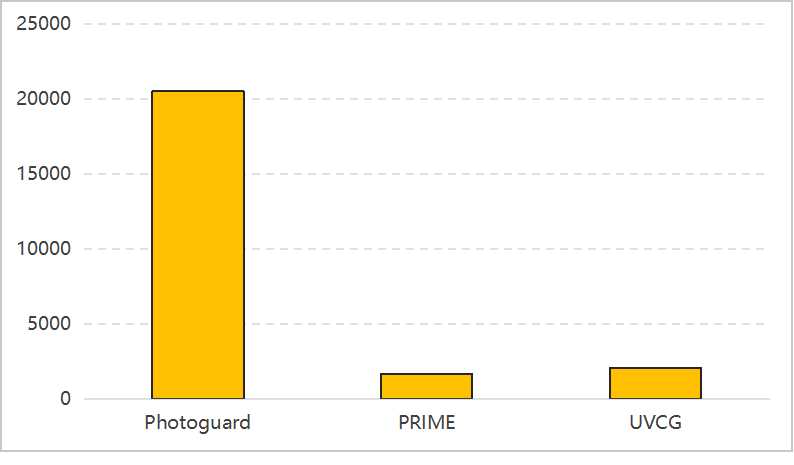}
  \caption{\textbf{GPU time consumption.} Protecting a frame video on an NVIDIA RTX A6000 takes 20,500 seconds with Photoguard, 1,700 seconds with PRIME\cite{prime}, and 2,100 seconds with UVCG.}
  \label{fig:cost}
\end{figure}

\noindent \textbf{Resource Consumption.} While ensuring effective video protection, it is equally important to account for the method’s resource consumption. PRIME\cite{prime} incorporates two strategies—fast convergence searching and early-stage stopping—to reduce GPU time
usage. Figure \ref{fig:cost} compares the GPU runtime required by various immunization methods to protect a 40-frame video. UVCG requires only 8.3\% of Photoguard’s GPU runtime, achieving the same level of efficiency as PRIME, which employs two resource-saving strategies. Additionally, UVCG requires 17 GB of GPU memory, making it suitable for consumer-grade GPUs, such as the RTX 3090.

\begin{table*}
    \centering
    \resizebox{\textwidth}{!}{%
        \begin{tabular}{ccccccccc}
            \toprule
            \multirow{2}{*}{\textbf{Editing pipeline}} & \multirow{2}{*}{\textbf{Immunization method}} 
            & \multicolumn{2}{c}{\textbf{Consistency score}} 
            & \multicolumn{4}{c}{\textbf{Similarity score}} \\ \cmidrule(lr){3-4} \cmidrule(lr){5-8}
            & & \textbf{Prompt consistency} & \textbf{Frame consistency} 
            & \textbf{SSIM↑} & \textbf{PSNR↑} & \textbf{LPIPS↓} & \textbf{VMAF↑} \\ 
            \midrule
            \multirow{4}{*}{\textbf{Tokenflow}} 
            & No immunization & 0.3231 & 0.9656 & \textbackslash & \textbackslash & \textbackslash & \textbackslash \\ 
            & Random noise immunization & 0.3176 & 0.9578 & 0.6376 & 22.72 & 0.3375 & 53.87 \\ 
            & UVCG (SD-v2.1) & \textbf{0.2887} & \textbf{0.9492} & \textbf{0.5118} & \textbf{19.58} & \textbf{0.4768} & \textbf{27.78} \\ 
            & UVCG (SD-v1.4) & 0.3048 & 0.9517 & 0.5747 & 21.15 & 0.4071 & 31.03 \\ 
            \midrule
            \multirow{4}{*}{\textbf{Text2Video-Zero}} 
            & No immunization & 0.3224 & 0.9471 & \textbackslash & \textbackslash & \textbackslash & \textbackslash \\ 
            & Random noise immunization & 0.3180 & 0.9451 & 0.7870 & 24.57 & 0.2394 & 71.94 \\ 
            & UVCG (SD-v2.1) & \textbf{0.3019} & \textbf{0.9326} & \textbf{0.5322} & \textbf{18.61} & \textbf{0.4717} & \textbf{32.20} \\ 
            & UVCG (SD-v1.4) & 0.3027 & 0.9309 & 0.5464 & 19.02 & 0.4522 & 34.72 \\ 
            \bottomrule
        \end{tabular}
    }
    \caption{Quantitative comparison of consistency and similarity scores for Tokenflow and Text2Video-zero under different immunization methods.Consistency scores evaluate the coherence between adjacent frames within a video and the alignment between the video content and the textual description;  Similarity scores measure the similarity between the edits of immune and non-immune videos. The arrows next to the metrics indicate an increase in video similarity. A lower similarity score suggests greater differences between the edits of the immune video and the original non-immune edited video.}
    \label{tab:metrics}
\end{table*}

\subsection{User study}
\begin{table}[ht]
    \centering
    \resizebox{\columnwidth}{!}{%
        \begin{tabular}{lcc}
            \toprule
            \textbf{Method} & \textbf{Video Quality} & \textbf{Immunization Success Rate} \\
            \midrule
            No immunization & 4.12 & \textbackslash \\
            Random noise immunization & 3.95 & 0.09 \\
            UVCG & \textbf{2.31} & \textbf{0.87} \\
            \bottomrule
        \end{tabular}
    }
    \caption{\textbf{User study results.} Comparison of video quality and immunization success rate under different methods.}
    \label{tab:user_study}
\end{table}
To further validate the effectiveness of UVCG, we conducted a user study. We selected four editing pipelines, each with five videos for editing. For each video, we provided three versions: the normally edited video, the video edited after baseline immunity, and the video edited after UVCG. Participants were asked to evaluate the videos based on three criteria: alignment with the textual description, naturalness, and inter-frame consistency, rating each video on a scale from 1 to 5, where 5 represents the highest quality. Additionally, participants indicated whether they found the two immunity methods effective or not. We collected 15 valid responses. The average evaluation results are presented in Table \ref{tab:user_study}. The following conclusions can be drawn from the results: 1)Existing editing pipelines can generate high-quality edited videos. 2)Random noise-based methods are ineffective in protecting videos. 3)UVCG significantly degrades video quality, providing robust protection. Specifically, the average video quality score dropped from 4.12 to 2.31 after applying UVCG, with an 87\% success rate in achieving immunity, confirming the effectiveness of the proposed method.
\section{Limitation}
\label{sec:limitation}
UVCG faces inherent challenges in targeted defense. Its protective effect may be less pronounced when simple edits are applied to videos with complex content. For example, if a video contains diverse elements such as animals, people, and landscapes, and the attacker performs straightforward semantic changes (e.g., day-to-night conversion), the protection may diminish (see Figure \ref{fig:result_Limitation} in Appendix). Nonetheless, it is important to note that such modifications can still be executed efficiently and at low cost using conventional video tuning tools, even without the use of AI-based editing models.
\section{Conclusion}
\label{sec:conclusion}
In this paper, we introduce a novel and generalizable method for safeguarding videos from malicious editing, called Universal Video Consistency Guard (UVCG). UVCG leverages the characteristics of video content by introducing continuous, imperceptible perturbations to "immunize" the video. The immunization disrupts editing process by forcing the encoder to map continuous inputs of protected video to misaligned continuous outputs. Extensive experiments have demonstrated the significant effectiveness, generalizability and efficiency of our method in safeguarding video content.
{
    \small
    \bibliographystyle{ieeenat_fullname}
    \bibliography{main}

\begin{thebibliography}{35}
\providecommand{\natexlab}[1]{#1}
\providecommand{\url}[1]{\texttt{#1}}
\expandafter\ifx\csname urlstyle\endcsname\relax
  \providecommand{\doi}[1]{doi: #1}\else
  \providecommand{\doi}{doi: \begingroup \urlstyle{rm}\Url}\fi

\bibitem[Bodur et~al.(2024)Bodur, Gundogdu, Bhattarai, Kim, Donoser, and Bazzani]{bodur2024iedit}
Rumeysa Bodur, Erhan Gundogdu, Binod Bhattarai, Tae-Kyun Kim, Michael Donoser, and Loris Bazzani.
\newblock iedit: Localised text-guided image editing with weak supervision.
\newblock In \emph{ICCV}, pages 7426--7435, 2024.

\bibitem[Brooks et~al.(2023{\natexlab{a}})Brooks, Holynski, and Efros]{brooks2023instructpix2pix}
Tim Brooks, Aleksander Holynski, and Alexei~A Efros.
\newblock Instructpix2pix: Learning to follow image editing instructions.
\newblock In \emph{ICCV}, pages 18392--18402, 2023{\natexlab{a}}.

\bibitem[Brooks et~al.(2023{\natexlab{b}})Brooks, Holynski, and Efros]{instructpix2pix}
Tim Brooks, Aleksander Holynski, and Alexei~A Efros.
\newblock Instructpix2pix: Learning to follow image editing instructions.
\newblock In \emph{CVPR}, pages 18392--18402, 2023{\natexlab{b}}.

\bibitem[Esser et~al.(2023)Esser, Chiu, Atighehchian, Granskog, and Germanidis]{gen-1}
Patrick Esser, Johnathan Chiu, Parmida Atighehchian, Jonathan Granskog, and Anastasis Germanidis.
\newblock Structure and content-guided video synthesis with diffusion models.
\newblock In \emph{ICCV}, pages 7346--7356, 2023.

\bibitem[Geyer et~al.(2023)Geyer, Bar-Tal, Bagon, and Dekel]{tokenflow}
Michal Geyer, Omer Bar-Tal, Shai Bagon, and Tali Dekel.
\newblock Tokenflow: Consistent diffusion features for consistent video editing.
\newblock \emph{arXiv preprint arXiv:2307.10373}, 2023.

\bibitem[Goodfellow et~al.(2014{\natexlab{a}})Goodfellow, Pouget-Abadie, Mirza, Xu, Warde-Farley, Ozair, Courville, and Bengio]{goodfellow2014generative}
Ian Goodfellow, Jean Pouget-Abadie, Mehdi Mirza, Bing Xu, David Warde-Farley, Sherjil Ozair, Aaron Courville, and Yoshua Bengio.
\newblock Generative adversarial nets.
\newblock \emph{NeurIPS}, 27, 2014{\natexlab{a}}.

\bibitem[Goodfellow et~al.(2014{\natexlab{b}})Goodfellow, Shlens, and Szegedy]{goodfellow2014explaining}
Ian~J Goodfellow, Jonathon Shlens, and Christian Szegedy.
\newblock Explaining and harnessing adversarial examples.
\newblock \emph{arXiv preprint arXiv:1412.6572}, 2014{\natexlab{b}}.

\bibitem[Gu(2024)]{gu2024responsible}
Jindong Gu.
\newblock Responsible generative ai: What to generate and what not.
\newblock \emph{arXiv preprint arXiv:2404.05783}, 2024.

\bibitem[He et~al.(2024)He, Zhu, Chen, Wang, and Gao]{he2024diff}
Xiao He, Mingrui Zhu, Dongxin Chen, Nannan Wang, and Xinbo Gao.
\newblock Diff-privacy: Diffusion-based face privacy protection.
\newblock \emph{IEEE TCSVT}, 2024.

\bibitem[Ho et~al.(2020)Ho, Jain, and Abbeel]{DiffusionModel}
Jonathan Ho, Ajay Jain, and Pieter Abbeel.
\newblock Denoising diffusion probabilistic models.
\newblock \emph{NeurIPS}, 33:\penalty0 6840--6851, 2020.

\bibitem[Kang et~al.(2024)Kang, Song, and Li]{diffattack}
Mintong Kang, Dawn Song, and Bo Li.
\newblock Diffattack: Evasion attacks against diffusion-based adversarial purification.
\newblock \emph{NeurIPS}, 36, 2024.

\bibitem[Khachatryan et~al.(2023)Khachatryan, Movsisyan, Tadevosyan, Henschel, Wang, Navasardyan, and Shi]{text2video}
Levon Khachatryan, Andranik Movsisyan, Vahram Tadevosyan, Roberto Henschel, Zhangyang Wang, Shant Navasardyan, and Humphrey Shi.
\newblock Text2video-zero: Text-to-image diffusion models are zero-shot video generators.
\newblock In \emph{ICCV}, pages 15954--15964, 2023.

\bibitem[Kingma(2013)]{vae}
Diederik~P Kingma.
\newblock Auto-encoding variational bayes.
\newblock \emph{arXiv preprint arXiv:1312.6114}, 2013.

\bibitem[Li et~al.(2024)Li, Yang, Zhang, and Zhang]{prime}
Guanlin Li, Shuai Yang, Jie Zhang, and Tianwei Zhang.
\newblock Prime: Protect your videos from malicious editing.
\newblock \emph{arXiv preprint arXiv:2402.01239}, 2024.

\bibitem[Li et~al.(2023)Li, Li, Savarese, and Hoi]{blip}
Junnan Li, Dongxu Li, Silvio Savarese, and Steven Hoi.
\newblock Blip-2: Bootstrapping language-image pre-training with frozen image encoders and large language models.
\newblock In \emph{International conference on machine learning}, pages 19730--19742, 2023.

\bibitem[Liang and Wu(2023)]{mist}
Chumeng Liang and Xiaoyu Wu.
\newblock Mist: Towards improved adversarial examples for diffusion models.
\newblock \emph{arXiv preprint arXiv:2305.12683}, 2023.

\bibitem[Liang et~al.(2023)Liang, Wu, Hua, Zhang, Xue, Song, Xue, Ma, and Guan]{liang2023adversarial}
Chumeng Liang, Xiaoyu Wu, Yang Hua, Jiaru Zhang, Yiming Xue, Tao Song, Zhengui Xue, Ruhui Ma, and Haibing Guan.
\newblock Adversarial example does good: Preventing painting imitation from diffusion models via adversarial examples.
\newblock \emph{arXiv preprint arXiv:2302.04578}, 2023.

\bibitem[Liao et~al.(2023)Liao, Chen, Yi, Yang, Wu, and Cao]{liao2023inter}
Junpei Liao, Zhikai Chen, Liang Yi, Wenyuan Yang, Baoyuan Wu, and Xiaochun Cao.
\newblock Inter-frame accelerate attack against video interpolation models.
\newblock \emph{arXiv preprint arXiv:2305.06540}, 2023.

\bibitem[Liu et~al.(2024)Liu, Zhang, Li, Lin, and Jia]{video-p2p}
Shaoteng Liu, Yuechen Zhang, Wenbo Li, Zhe Lin, and Jiaya Jia.
\newblock Video-p2p: Video editing with cross-attention control.
\newblock In \emph{Proceedings of the IEEE/CVF Conference on Computer Vision and Pattern Recognition}, pages 8599--8608, 2024.

\bibitem[Liu et~al.(2016)Liu, Chen, Liu, and Song]{liu2016delving}
Yanpei Liu, Xinyun Chen, Chang Liu, and Dawn Song.
\newblock Delving into transferable adversarial examples and black-box attacks.
\newblock \emph{arXiv preprint arXiv:1611.02770}, 2016.

\bibitem[Madry(2017)]{pgd}
Aleksander Madry.
\newblock Towards deep learning models resistant to adversarial attacks.
\newblock \emph{arXiv preprint arXiv:1706.06083}, 2017.

\bibitem[Papernot et~al.(2016)Papernot, McDaniel, and Goodfellow]{papernot2016transferability}
Nicolas Papernot, Patrick McDaniel, and Ian Goodfellow.
\newblock Transferability in machine learning: from phenomena to black-box attacks using adversarial samples.
\newblock \emph{arXiv preprint arXiv:1605.07277}, 2016.

\bibitem[Pont-Tuset et~al.(2017)Pont-Tuset, Perazzi, Caelles, Arbel{\'a}ez, Sorkine-Hornung, and Van~Gool]{davis}
Jordi Pont-Tuset, Federico Perazzi, Sergi Caelles, Pablo Arbel{\'a}ez, Alex Sorkine-Hornung, and Luc Van~Gool.
\newblock The 2017 davis challenge on video object segmentation.
\newblock \emph{arXiv preprint arXiv:1704.00675}, 2017.

\bibitem[Qi et~al.(2023)Qi, Cun, Zhang, Lei, Wang, Shan, and Chen]{fatezero}
Chenyang Qi, Xiaodong Cun, Yong Zhang, Chenyang Lei, Xintao Wang, Ying Shan, and Qifeng Chen.
\newblock Fatezero: Fusing attentions for zero-shot text-based video editing.
\newblock In \emph{ICCV}, pages 15932--15942, 2023.

\bibitem[Radford et~al.(2021)Radford, Kim, Hallacy, Ramesh, Goh, Agarwal, Sastry, Askell, Mishkin, Clark, et~al.]{clip}
Alec Radford, Jong~Wook Kim, Chris Hallacy, Aditya Ramesh, Gabriel Goh, Sandhini Agarwal, Girish Sastry, Amanda Askell, Pamela Mishkin, Jack Clark, et~al.
\newblock Learning transferable visual models from natural language supervision.
\newblock In \emph{International conference on machine learning}, pages 8748--8763, 2021.

\bibitem[Rombach et~al.(2022)Rombach, Blattmann, Lorenz, Esser, and Ommer]{LDM}
Robin Rombach, Andreas Blattmann, Dominik Lorenz, Patrick Esser, and Bj{\"o}rn Ommer.
\newblock High-resolution image synthesis with latent diffusion models.
\newblock In \emph{CVPR}, pages 10684--10695, 2022.

\bibitem[Salman et~al.(2023)Salman, Khaddaj, Leclerc, Ilyas, and Madry]{photoguard}
Hadi Salman, Alaa Khaddaj, Guillaume Leclerc, Andrew Ilyas, and Aleksander Madry.
\newblock Raising the cost of malicious ai-powered image editing.
\newblock \emph{arXiv preprint arXiv:2302.06588}, 2023.

\bibitem[Shan et~al.(2023)Shan, Cryan, Wenger, Zheng, Hanocka, and Zhao]{glaze}
Shawn Shan, Jenna Cryan, Emily Wenger, Haitao Zheng, Rana Hanocka, and Ben~Y Zhao.
\newblock Glaze: Protecting artists from style mimicry by $\{$Text-to-Image$\}$ models.
\newblock In \emph{USENIX}, pages 2187--2204, 2023.

\bibitem[Sheikh and Bovik(2006)]{vmaf}
Hamid~R Sheikh and Alan~C Bovik.
\newblock Image information and visual quality.
\newblock \emph{ICIP}, 15\penalty0 (2):\penalty0 430--444, 2006.

\bibitem[Wang et~al.(2004)Wang, Bovik, Sheikh, and Simoncelli]{ssim}
Zhou Wang, Alan~C Bovik, Hamid~R Sheikh, and Eero~P Simoncelli.
\newblock Image quality assessment: from error visibility to structural similarity.
\newblock \emph{ICIP}, 13\penalty0 (4):\penalty0 600--612, 2004.

\bibitem[Wu et~al.(2023)Wu, Ge, Wang, Lei, Gu, Shi, Hsu, Shan, Qie, and Shou]{tuneavideo}
Jay~Zhangjie Wu, Yixiao Ge, Xintao Wang, Stan~Weixian Lei, Yuchao Gu, Yufei Shi, Wynne Hsu, Ying Shan, Xiaohu Qie, and Mike~Zheng Shou.
\newblock Tune-a-video: One-shot tuning of image diffusion models for text-to-video generation.
\newblock In \emph{ICCV}, pages 7623--7633, 2023.

\bibitem[Yu et~al.(2024)Yu, Zhang, Xu, and Zhang]{yu2024cross}
Jiwen Yu, Xuanyu Zhang, Youmin Xu, and Jian Zhang.
\newblock Cross: Diffusion model makes controllable, robust and secure image steganography.
\newblock \emph{NeurIPS}, 36, 2024.

\bibitem[Zhang et~al.(2018)Zhang, Isola, Efros, Shechtman, and Wang]{lpips}
Richard Zhang, Phillip Isola, Alexei~A Efros, Eli Shechtman, and Oliver Wang.
\newblock The unreasonable effectiveness of deep features as a perceptual metric.
\newblock In \emph{CVPR}, pages 586--595, 2018.

\bibitem[Zhang et~al.(2023)Zhang, Han, Ghosh, Metaxas, and Ren]{zhang2023sine}
Zhixing Zhang, Ligong Han, Arnab Ghosh, Dimitris~N Metaxas, and Jian Ren.
\newblock Sine: Single image editing with text-to-image diffusion models.
\newblock In \emph{ICCV}, pages 6027--6037, 2023.

\bibitem[Zheng et~al.(2023)Zheng, Liang, Wu, and Liu]{zheng2023understanding}
Boyang Zheng, Chumeng Liang, Xiaoyu Wu, and Yan Liu.
\newblock Understanding and improving adversarial attacks on latent diffusion model.
\newblock \emph{arXiv preprint arXiv:2310.04687}, 2023.

\end{thebibliography}
}
\clearpage
\setcounter{page}{1}
\maketitlesupplementary
\section{Latent Diffusion Model}
\label{sec:Latent Diffusion Model}
Latent Diffusion Models (LDMs)\cite{LDM} represent a novel class of generative models that integrate the principles of diffusion processes\cite{DiffusionModel} and variational autoencoders (VAEs)\cite{vae}. Unlike traditional diffusion models that operate directly on high-dimensional data, LDMs map data into a lower-dimensional latent space where the diffusion process unfolds, significantly reducing the computational complexity while preserving the expressive power of the model.

Given a data distribution \( \mathbf{x} \sim p(\mathbf{x}) \), LDMs encode the data into a latent representation \( \mathbf{z} \in \mathbb{R}^d \) using an encoder \( \mathcal{E} \), where \( \mathbf{z} = \mathcal{E}(\mathbf{x}) \). The latent space preserves essential characteristics of \( \mathbf{x} \) while reduces dimensionality, such that \( d \ll \text{dim}(\mathbf{x}) \). The latent representation then undergoes a diffusion process, during which noise is incrementally added to \( \mathbf{z} \) over time, resulting in a noise distribution \( p(\mathbf{z}_T) \).
The forward diffusion process can be described as:
\[
\mathbf{z}_{t+1} = \sqrt{\alpha_t} \mathbf{z}_t + \sqrt{1 - \alpha_t} \epsilon_t, \quad \epsilon_t \sim \mathcal{N}(0, \mathbf{I}),
\]
where \( t \) represents the inversion step, and \( \alpha_t \) controls the noise injection rate. The model learns to reverse this process by progressively denoising the latent variables using a trained neural network \( f_\theta \), such that:
\[
\tilde{\mathbf{z}}_t = f_\theta(\mathbf{z}_{t+1}, t).
\]
After reversing the diffusion, the denoised latent variable \( \tilde{\mathbf{z}}_0 \) is decoded back into the data space via a decoder \( D \), reconstructing the sample \( \tilde{\mathbf{x}} = D(\tilde{\mathbf{z}}_0) \).

LDMs also support conditional generation, allowing the sampling process to be guided by external conditions (e.g., images or natural language). This is achieved by combining the latent representation \( \mathbf{z}_T \), obtained during the diffusion process, with the embedding of condition. The denoising network  \( f_\theta \) processes the combined representation for \( T \) steps, producing a modified latent variable \( \tilde{\mathbf{z}} \), which is then passed through the decoder \( D \) to generate a new output, similar to the unconditional case.

\section{Additional Results}
\label{sec:addition Results}
\begin{figure*}
  \centering
  \includegraphics[width=\linewidth]{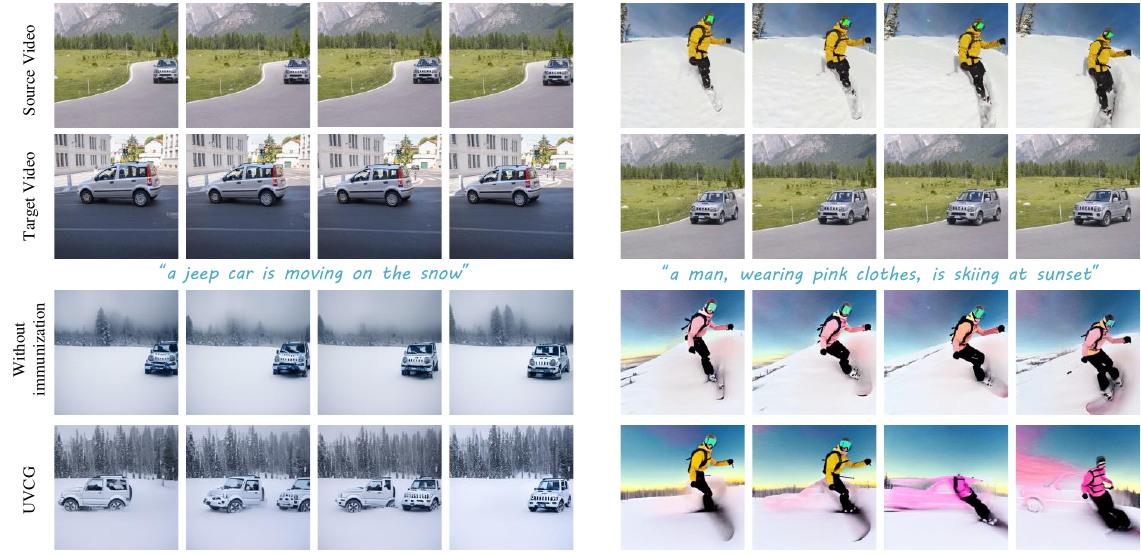}
  \caption{\textbf{Protection Effectiveness on Tune-A-Video\cite{tuneavideo}.} The base model employed for editing in Text2Video-zero is Stable Diffusion-v1.4. First row: the original video. Second row: the target video. Third row: editing results without immunization. Fourth row: editing results after applying UVCG with SD-v2.1 as the protection model.}
  \label{fig:result_TuneAVideo_01}
\end{figure*}

\begin{figure*}
  \centering
  \includegraphics[width=\linewidth]{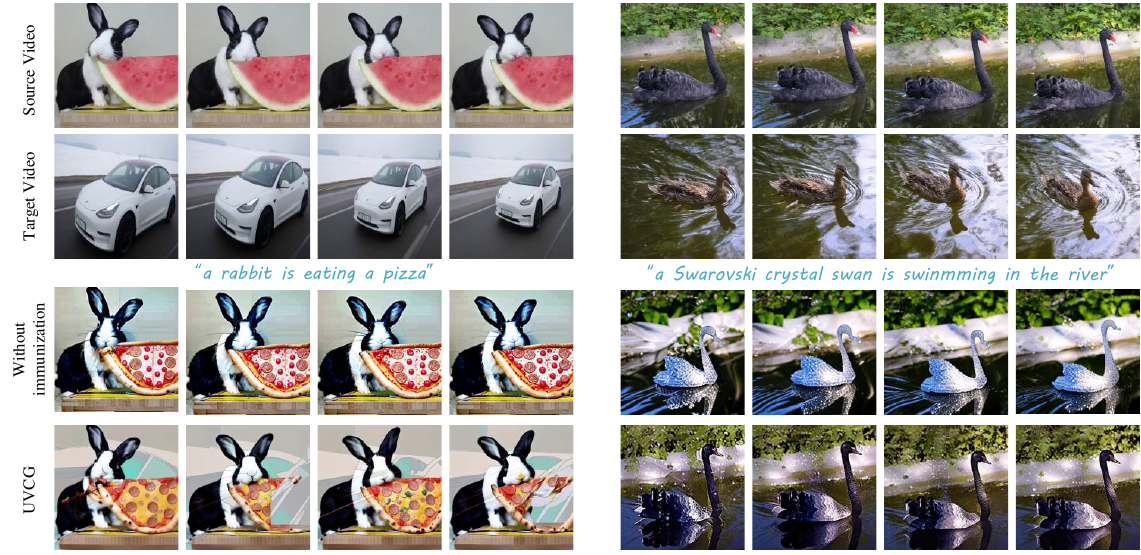}
  \caption{\textbf{Protection Effectiveness on Tune-A-Video\cite{tuneavideo}.} The base model employed for editing in Text2Video-zero is Stable Diffusion-v1.4. First row: the original video. Second row: the target video. Third row: editing results without immunization. Fourth row: editing results after applying UVCG with SD-v2.1 as the protection model.}
  \label{fig:result_TuneAVideo_02}
\end{figure*}

\begin{figure*}
  \centering
  \includegraphics[width=\linewidth]{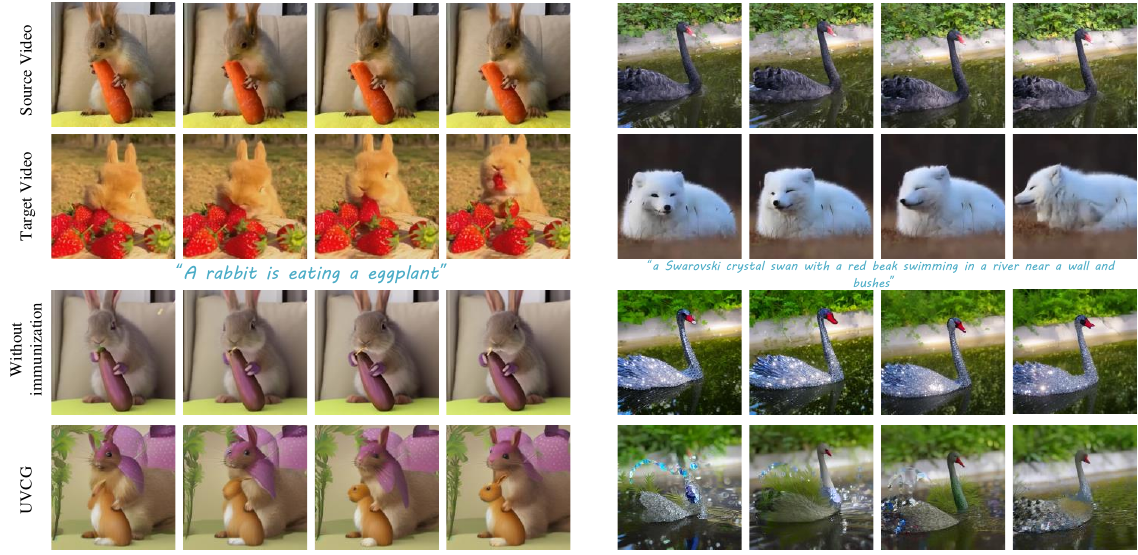}
  \caption{\textbf{Protection Effectiveness on Fatezero\cite{fatezero}.} The base model employed for editing in Text2Video-zero is Stable Diffusion-v1.4. First row: the original video. Second row: the target video. Third row: editing results without immunization. Fourth row: editing results after applying UVCG with SD-v2.1 as the protection model.}
  \label{fig:result_Fatezero_01}
\end{figure*}

\begin{figure*}
  \centering
  \includegraphics[width=\linewidth]{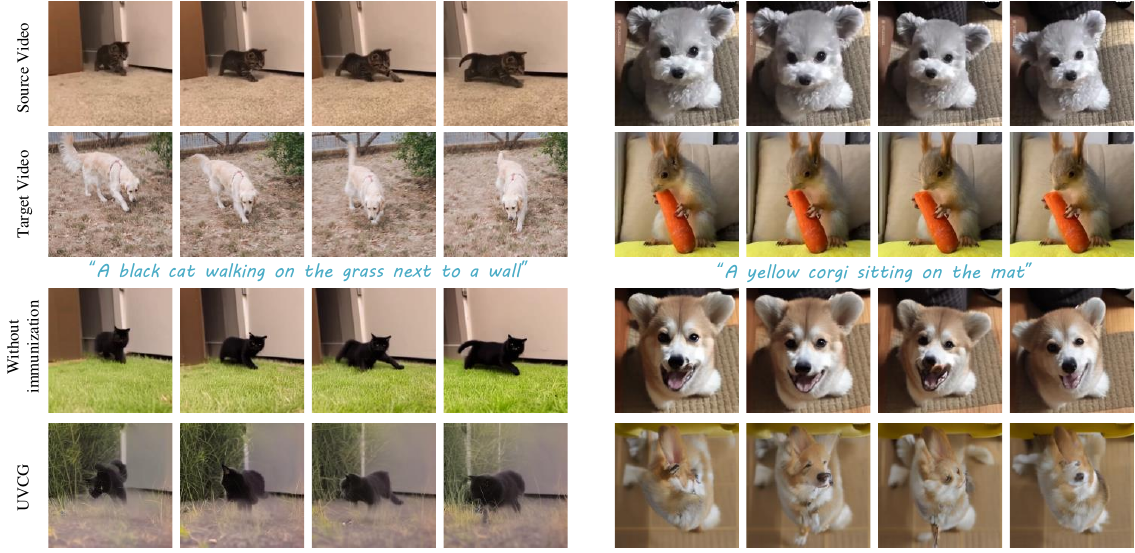}
  \caption{\textbf{Protection Effectiveness on Fatezero\cite{fatezero}.} The base model employed for editing in Text2Video-zero is Stable Diffusion-v1.4. First row: the original video. Second row: the target video. Third row: editing results without immunization. Fourth row: editing results after applying UVCG with SD-v2.1 as the protection model.}
  \label{fig:result_Fatezero_02}
\end{figure*}

\begin{figure*}
  \centering
  \includegraphics[width=\linewidth]{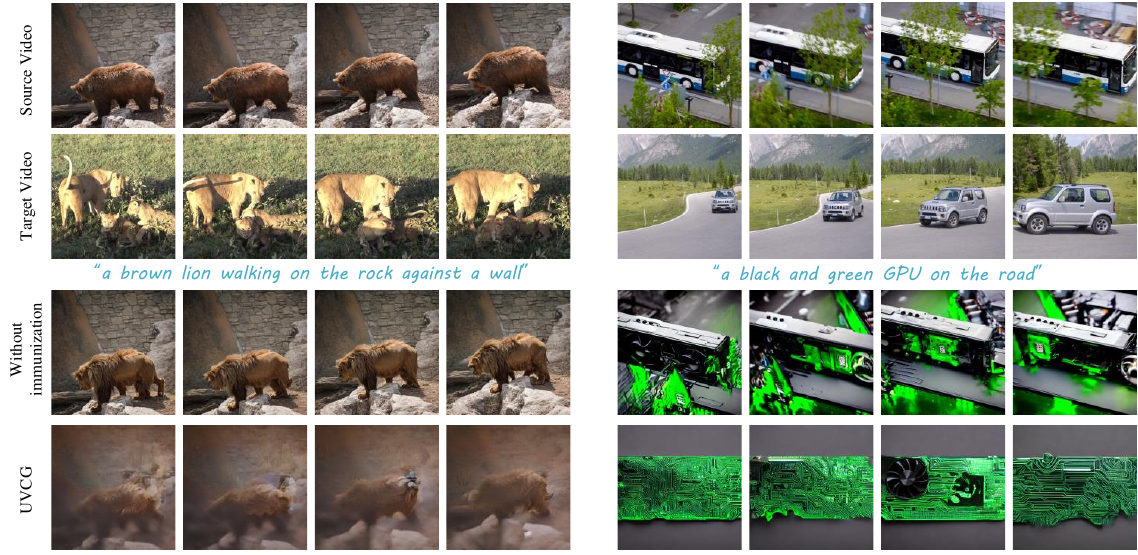}
  \caption{\textbf{Protection Effectiveness on Fatezero\cite{fatezero}.} The base model employed for editing in Text2Video-zero is Stable Diffusion-v1.4. First row: the original video. Second row: the target video. Third row: editing results without immunization. Fourth row: editing results after applying UVCG with SD-v2.1 as the protection model.}
  \label{fig:result_Fatezero_03}
\end{figure*}

\begin{figure*}
  \centering
  \includegraphics[width=\linewidth]{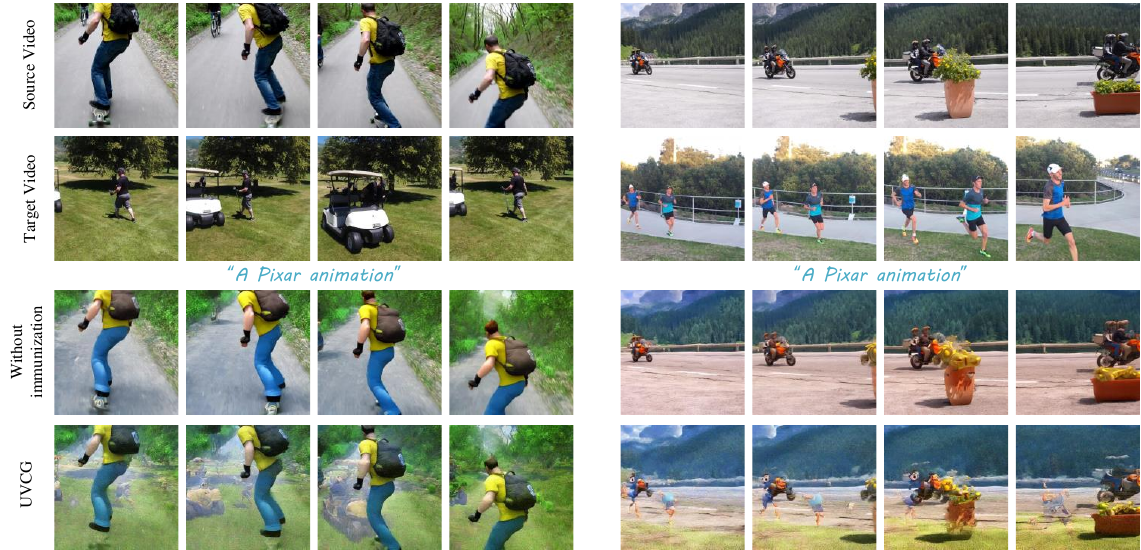}
  \caption{\textbf{Protection Effectiveness on Tokenflow\cite{tokenflow}.} The base model employed for editing in Text2Video-zero is Stable Diffusion-v2.1. First row: the original video. Second row: the target video. Third row: editing results without immunization. Fourth row: editing results after applying UVCG with SD-v1.4 as the protection model.}
  \label{fig:results_tokenflow_02}
\end{figure*}

\begin{figure*}
  \centering
  \includegraphics[width=\linewidth]{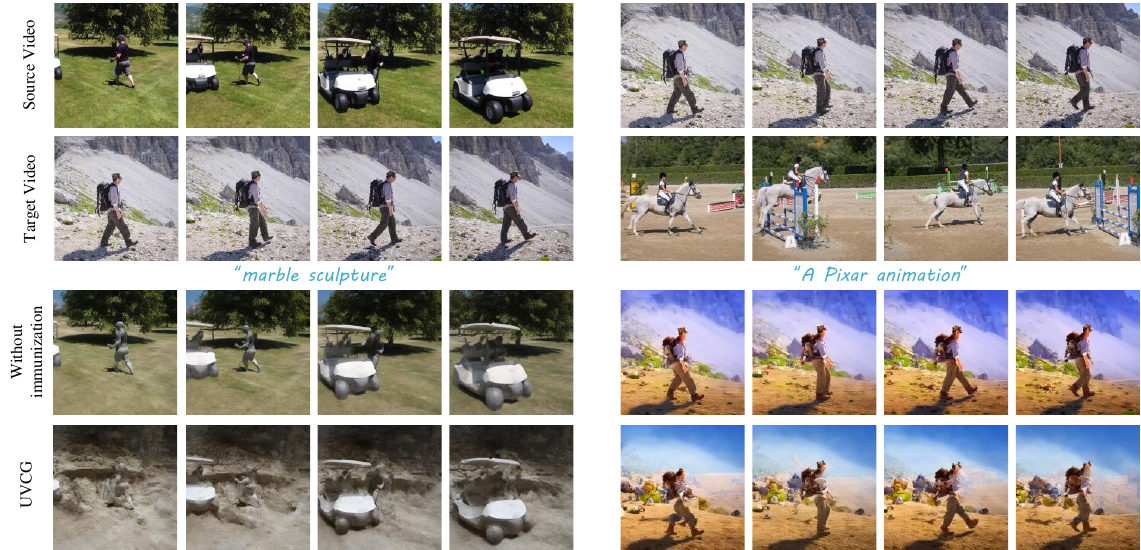}
  \caption{\textbf{Protection Effectiveness on Tokenflow\cite{tokenflow}.} The base model employed for editing in Text2Video-zero is Stable Diffusion-v2.1. First row: the original video. Second row: the target video. Third row: editing results without immunization. Fourth row: editing results after applying UVCG with SD-v1.4 as the protection model.}
  \label{fig:results_tokenflow_03}
\end{figure*}

\begin{figure*}
  \centering
  \includegraphics[width=\linewidth]{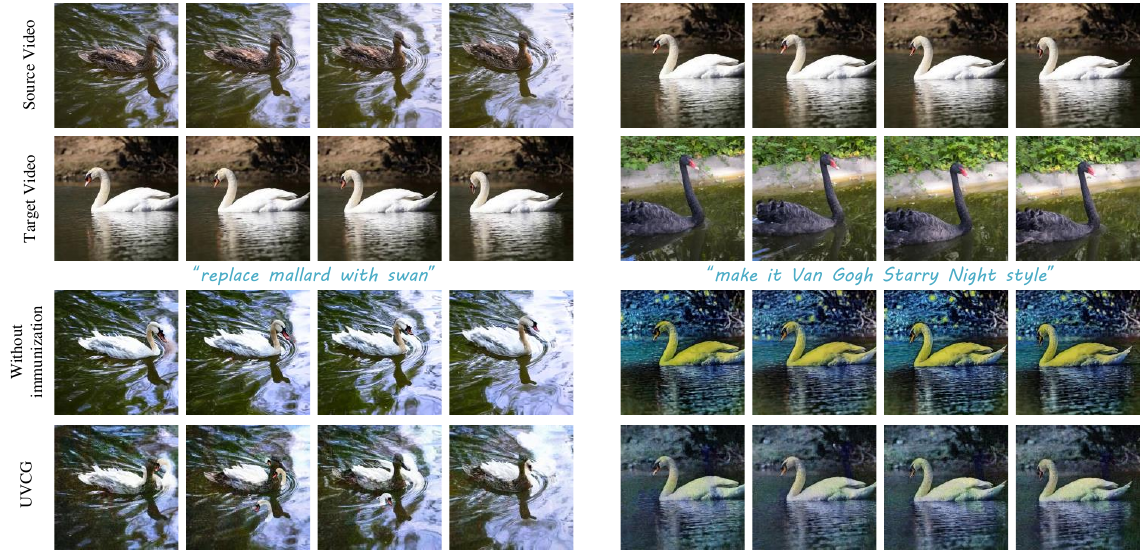}
  \caption{\textbf{Protection Effectiveness on Text2Video-zero\cite{text2video}.} The base model employed for editing in Text2Video-zero is Instruct-pix2pix\cite{instructpix2pix}. First row: the original video. Second row: the target video. Third row: editing results without immunization. Fourth row: editing results after applying UVCG with SD-v1.4 as the protection model.}
  \label{fig:results_text2video_02}
\end{figure*}

\begin{figure*}
  \centering
  \includegraphics[width=\linewidth]{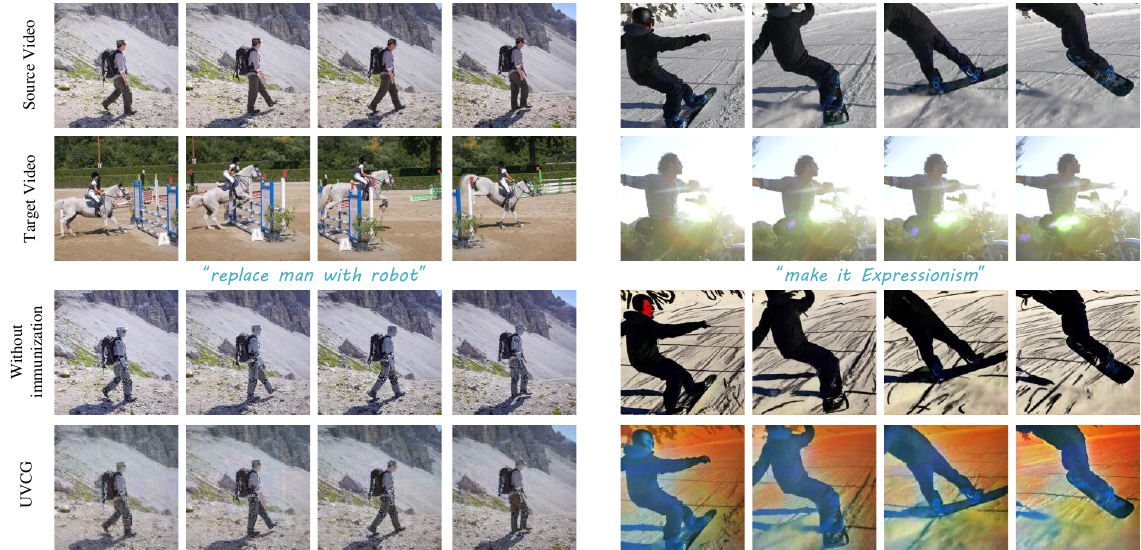}
  \caption{\textbf{Protection Effectiveness on Text2Video-zero\cite{text2video}.} The base model employed for editing in Text2Video-zero is Instruct-pix2pix\cite{instructpix2pix}. First row: the original video. Second row: the target video. Third row: editing results without immunization. Fourth row: editing results after applying UVCG with SD-v1.4 as the protection model.}
  \label{fig:results_text2video_03}
\end{figure*}

Figure \ref{fig:result_TuneAVideo_01} and figure \ref{fig:result_TuneAVideo_02} showcase examples of video protection results using our method on Tune-A-Video. Figure \ref{fig:result_Fatezero_01}, figure \ref{fig:result_Fatezero_02} and figure \ref{fig:result_Fatezero_03} showcase video protection results using our method on Fatezero. Figure \ref{fig:results_tokenflow_02} adn figure \ref{fig:results_tokenflow_03} provides additional examples of video protection results using our method on Tokenflow. Figure \ref{fig:results_text2video_02} and \ref{fig:results_text2video_03} provides additional examples of video protection results using our method on Text2Video. 
\begin{figure*}
  \centering
  \includegraphics[width=\linewidth]{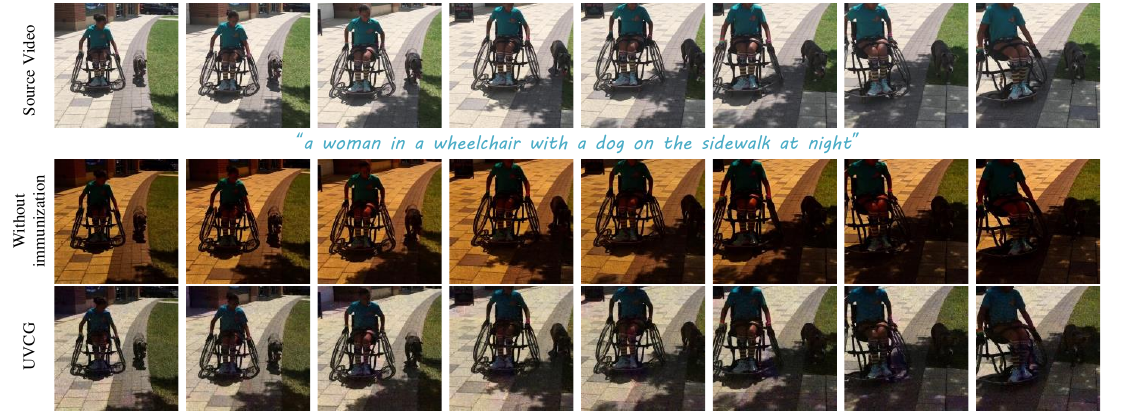}
  \caption{The protection effectiveness of UVCG on simple editing semantics (e.g., transitioning from day to night, which corresponds to reducing the brightness of the video).}
  \label{fig:result_Limitation}
\end{figure*}

% \input{sec/cvpr_template}
% \input{sec/2_formatting}
% \input{sec/3_finalcopy}


\end{document}